\documentclass{article}
\usepackage{spconf,amsmath,epsfig}
\usepackage{amsfonts,amssymb}
\usepackage{color, xcolor}

\let\OLDthebibliography\thebibliography
\renewcommand\thebibliography[1]{
  \OLDthebibliography{#1}
  \setlength{\parskip}{0pt}
  \setlength{\itemsep}{0pt plus 0.3ex}
}

\begin{document}\sloppy

% Example definitions.
% --------------------
\def\x{{\mathbf x}}
\def\L{{\cal L}}

% Title.
% ------
\title{Prompt-IML: Image Manipulation Localization with pre-trained foundation models through prompt tuning}
%
% Single address.
% ---------------
% \name{Anonymous ICME submission}
%Address and e-mail should NOT be added in the submission paper. They should be present only in the camera ready paper. 
% \address{}

\name{Xuntao Liu, Yuzhou Yang, Qichao Ying, Zhenxing Qian$^{\ast}$\thanks{$^{\ast}$ indicates the corresponding author. This work was supported by the National Natural Science Foundation of China under Grants U20B2051 and U1936214.}, Xinpeng Zhang and Sheng Li}
\address{School of Computer Science, Fudan University, China\\
\{22210240093@m., 22110240074@m., qcying20@, zxqian@, zhangxinpeng@, lisheng@\}fudan.edu.cn\\
}

\maketitle

\begin{abstract}
Deceptive images can be shared in seconds with social networking services, posing substantial risks. Tampering traces, such as boundary artifacts and high-frequency information, have been significantly emphasized by massive networks in the Image Manipulation Localization (IML) field. However, they are prone to image post-processing operations, which limit the generalization and robustness of existing methods. 
We present a novel Prompt-IML framework. We observe that humans tend to discern the authenticity of an image based on both semantic and high-frequency information, inspired by which, the proposed framework leverages rich semantic knowledge from pre-trained visual foundation models to assist IML.
We are the first to design a framework that utilizes visual foundation models specially for the IML task.
Moreover, we design a Feature Alignment and Fusion module to align and fuse features of semantic features with high-frequency features, which aims at locating tampered regions from multiple perspectives.  Experimental results demonstrate that our model can achieve better performance on eight typical fake image datasets and outstanding robustness.
\end{abstract}
\section{Introduction}
\label{sec:intro}
The commonly encountered image processing techniques are copy-move, splicing, and inpainting, all of which have the capability to alter the original semantic content of images. Meanwhile, the rapid advancement of image editing tools has substantially reduced the difficulty and cost associated with producing deceptive images. Consequently, there is a pressing need to accurately locate the manipulated regions in deceptive images.

The Image Manipulation Localization (IML) task is to granularly locate tampered regions within images. With the advancement of deep learning, researchers are attempting to establish massive manipulation localization networks~\cite{PSCC}. 
Many existing methods revolve around specific tampering traces, seeking the optimal feature representation through carefully designed network architectures~\cite{RRU-Net,wu2019mantra,MFCN}. However, tampering traces are prone to image post-processing operations~\cite{DRAW,MVSS-Net++}. It's a contributing factor to the limited robustness and generalization of the aforementioned methods.

\begin{figure}[!t]
  \centering
  \includegraphics[width=0.49\textwidth]{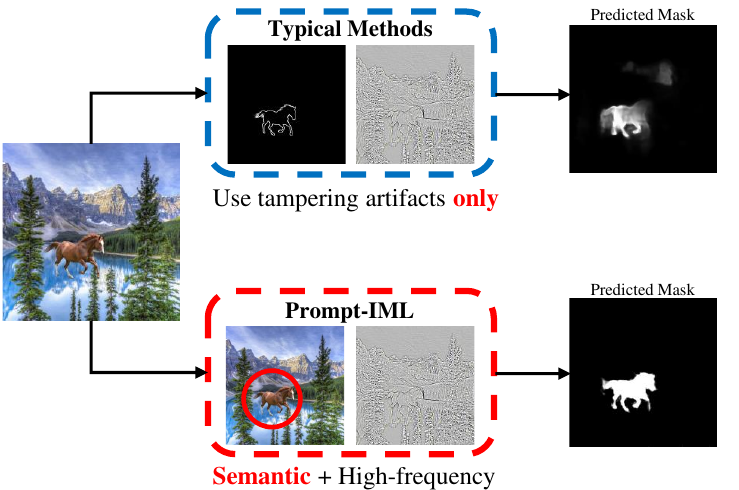} 
  \caption{Prompt-IML incorporates semantic information with high-frequency information to improve the performance. Semantic information may notice specific objects (circled red region) to assist the manipulation localization.} 
  \label{intro}
\end{figure}

% Moreover, we notice that there are slight differences between how humans and machines identify fake images. 
Moreover, we notice that existing methods ignore a key element of images to achieve generalizability and robustness, that is the semantic information~\cite{MVSS-Net++}.
Humans naturally observe the coherence of the semantic information within the picture to identify fake images. 
Semantic information plays a principal part in many computer vision tasks~\cite{Mask2Former}.
We believe that it is not exceptional in IML. 
Compared with features related to tampering traces, semantic features are more robust to image post-processing.
Therefore, employing the semantic features of images as another supplement for judgment will assist the task of IML. 
However, training a network with rich semantic knowledge using limited available datasets is challenging~\cite{MAE}. Besides, typical methods often utilize high-frequency features from images to identify manipulations, which brings new challenges of aligning semantic features with them~\cite{CMX}.

\begin{figure*}[!t]
  \centering
  \includegraphics[width=0.99\textwidth]{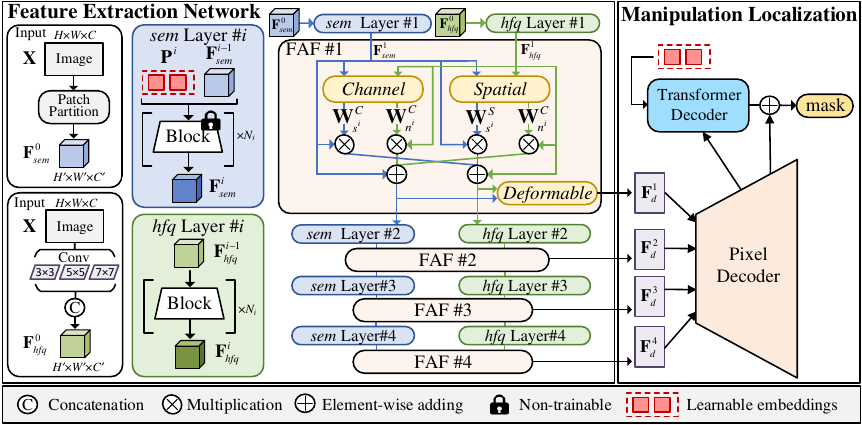} 
  \caption{Architecture overview. The \textit{Channel}, \textit{Spatial}, \textit{Deformable} represents the procedure of Eq.~\ref{channel_att}, Eq.~\ref{spatial_att}, and Eq.~\ref{deform_att}.} 
  \label{architecture}
\end{figure*}

% On the other hand, fine-tuning a pre-trained network to utilize semantic knowledge for IML is troublesome as it may damage the performance of learned knowledge~\cite{}. 

To overcome the aforementioned challenges, we exploit pre-trained visual foundation models to acquire semantic features of images through prompt tuning.
Fig.~\ref{intro} exhibits the difference between the proposed method and typical methods.
We propose prompt-IML that actively utilizes semantic information along with high-frequency information for manipulation localization.
Specifically, We use BayarConv to extract the high-frequency features of images and feed them into subsequent networks for further processing. Simultaneously, we raise a semantic feature extraction network, which adheres to the architectural design of visual foundation models and is initialized with pre-trained weights. During training, we freeze it and attach several learnable prompt embeddings to image token sequences to adjust the semantic features. Then, we facilitate interaction between semantic features and high-frequency features through a designed Feature Alignment and Fusion (FAF) module, which involves multiple attention mechanisms to enhance features and locate tampered regions from multiple perspectives.

Our main contributions are summarized as follows:
\begin{itemize}
\item We are the first to design a framework that utilizes visual foundation models specially for the IML task. Incorporating semantic information with high-frequency information for discernment aligns more with the logic circuit of human judgment regarding image veracity.
\item We propose an FAF module, that enables adapting visual foundation models to IML tasks through prompt tuning. The proposed FAF module involves multiple attention mechanisms to align and fuse semantic features with high-frequency features. 
\item Experiments on eight datasets demonstrate the generalizability of the proposed framework. Extensive experiments prove the robustness of the framework against image post-processing operations. 
\end{itemize}

\section{Related Works}
\textbf{Image Manipulation Localization}. With the advancement of deep learning, researchers have embarked on efforts to establish end-to-end manipulation localization networks. 
% HPFCN uses three learnable high-pass filters to extract noise features of images and then locate tampered regions through noise inconsistency. SPAN models the relationship between image patches at multiple scales by constructing a pyramid of local self-attention blocks. 
MVSSNet++\cite{MVSS-Net++} integrates multi-scale features, contour features, and high-frequency features of images for feature extraction and utilizes spatial-channel attention for enhanced feature fusion. PSCC-Net\cite{PSCC} proposes a progressive spatial-channel attention module, utilizing multi-scale features and dense cross-connections to generate tampering masks of various granularity. These works involve the meticulous design of network architectures to acquire more optimal feature representations regarding tampering traces. Although these methods have achieved decent performance in the IML task, the choice of feature representation for tampering traces still significantly impacts the model's generalization and robustness.

\noindent\textbf{Tuning Visual Foundation Models}. Compared to fine-tuning, prompt tuning is an efficient, low-cost way of adapting an AI foundation model to new downstream tasks without retraining the model and updating its weights. This technique was first used in NLP, and VPT\cite{VPT} is an efficient way to adapt it for the visual domain. Recently, EVP\cite{EVP} achieved granular manipulated region localization by adjusting the embedding representation of images and incorporating high-frequency information. They attempt to adapt the pre-trained model through prompt tuning to various downstream tasks, including IML. However, due to the simple feature fusion design, their model performs poorly.

\section{Proposed Method}
% In this section, we elaborate on our method for IML. We first introduce the overall framework, then we dive into the details of the proposed framework. Finally, we illustrate the prompt tuning method for training.

\subsection{Approach Overview}
Fig.~\ref{architecture} illustrates the architecture of the proposed prompt-IML. The complete pipeline consists of two phases, i.e., Feature Extraction Network (FEN) and Manipulation Localization Network (MLN).
The FEN comprises two parallel branches: one extracts semantic features, and the other focuses on extracting high-frequency features. Given the differences between them, we employ a carefully designed FAF module to fuse features. This module primarily utilizes various attention mechanisms to facilitate interaction between the features. The multi-scale features outputted during the FEN stage are ultimately fed into the MLN. The MLN aggregates feature information through layer-wise up-sampling and outputs the final prediction results.

\subsection{Feature Extraction Network}
\textbf{Dual-Branch Architecture}. We use two branches to extract multiple features of images, and both share the same structure based on Swin-Transformer~\cite{swin}. The semantic branch is initialized with pre-trained weights and remains untrained during training to preserve the optimal semantic representations. Specifically, let $\mathbf{X}\in \mathbb{R}^{H\times W\times C}$ be the input image. For the semantic branch, we get the input $\mathbf F^0_{sem}\in \mathbb{R}^{(H'\times W')\times C'}$ through partitioning the image into specified-sized patches:
\begin{equation}
    \mathbf F^0_{sem} = \operatorname{Norm}(\operatorname{Conv}(\mathbf X)) + \mathbf F_{PE}, 
\end{equation}
where $H'\times W'$ represents the number of the patches, $\mathbf F_{PE}$ is a learnable positional embedding. For the high-frequency branch, we employ a set of BayarConv with kernels varying sizes, which prevent information loss caused by a fixed-size receptive field, to get the input $\mathbf F^0_{\textit{hfq}}\in \mathbb{R}^{(H'\times W')\times C'}$:
\begin{equation}
\mathbf F^0_{\textit{hfq}} = \operatorname{Concat}(\{\operatorname{BayarConv_{i\times i}(\mathbf X)},\ i=3,5,7\}), 
\end{equation}
where $i$ symbolizes the kernel size.

% To comprehensively consider both global and local information and mitigate information loss\cite{}, multi-scale features are generated for subsequent tasks. Specifically, the branch comprises four layers, each of which consists of several blocks. Let the output of the $j$-th block in the $i$-th layer be $\mathbf F_{i,j} \in \mathbb{R}^{(H_i\times W_i)\times C_i}$, and the $i$-th layer's output be $\mathbf F_i$, equivalent to the last block's output. The forward propagation process in each block can be described below:
To comprehensively consider both global and local information and mitigate information loss\cite{MVSS-Net++}, multi-scale features $\mathbf F^{i}_{sem},\mathbf F^{i}_{\textit{sem}} \in \mathbb{R}^{(H_i\times W_i)\times C_i}, i=1,...,4$ are generated for the subsequent procedure in each branch. Specifically, each branch comprises four layers, namely \textit{sem} layer and \textit{hfq} layer, each of which consists of several blocks.
 The forward propagation process in each block can be described below:
\begin{equation}
\begin{split}
\mathbf z^{i,j} &= \operatorname{SelfAttn}(\operatorname{LN}(\mathbf F^{i,j-1}))+\mathbf F^{i,j-1}, \\
\mathbf F^{i,j} &= \operatorname{MLP}(\operatorname{LN}(\mathbf z^{i,j}))+\mathbf z^{i,j},
\end{split}
\end{equation}
where $j=1,...,N_i, \mathbf F^{i,0}=\mathbf F^i$, $\operatorname{LN}(\cdot)$ denotes layer normalization, and $\mathbf F^{i,j}$ denotes the output of the $i$-th layer and $j$-th block. 
% In each branch, we have $N_i=\{2,2,18,2\}, i=1,2,3,4$.

\begin{figure*}[!t]
  \centering
  \includegraphics[width=0.99\textwidth]{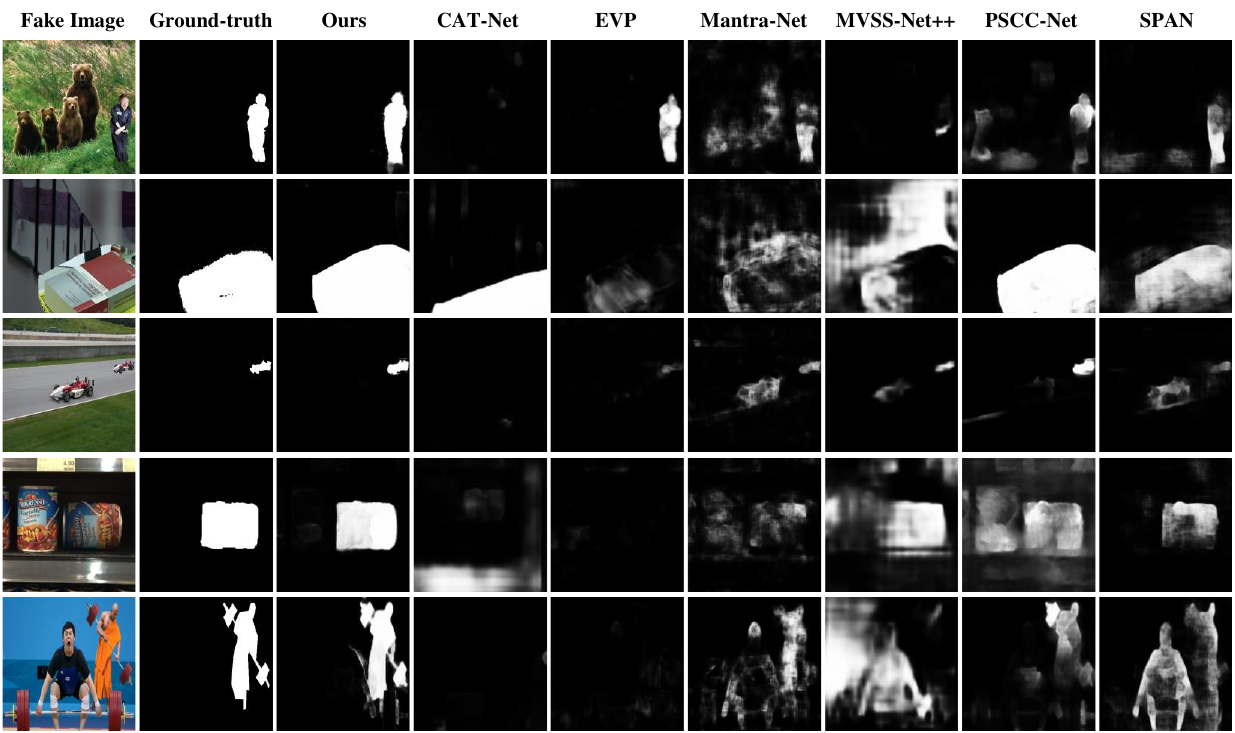} 
  \caption{Manipulation localization results on images originating from multiple datasets. The 3-rd column represents the results of our method, while columns 4 to 10 depict the results of another six SOTA methods.} 
  \label{lr}
\end{figure*}

\noindent\textbf{Feature Alignment and Fusion Module}. 
Attention mechanisms are widely used to enhance features in IML task~\cite{MVSS}. We propose a FAF module for better alignment and fusion of multiple features, which consists of \textit{channel attention}, \textit{spatial attention}, and \textit{deformable attention}. 
% To simplify the description, we use $\mathbf F_{s^i}\in \mathbb{R}^{(H_i\times W_i)\times C_i}$ and $\mathbf F_{h^i}\in \mathbb{R}^{(H_i\times W_i)\times C_i}$ to represent the $i$-th layer's output of semantic features and high-frequency features, respectively. 
First, we employ average pooling operation to reduce features, denoted by overline. Then, they are concatenated on dimension $C_i$, which is denoted by $[\cdot]$, and fed into an MLP to generate corresponding channel-attention vectors $\mathbf W^C_{s^i}, \mathbf W^C_{h^i}\in \mathbb{R}^{C_i}$, the above procedure can be formulated as:
\begin{equation}
\label{channel_att}
\begin{split}
% \mathbf Z_{s^i}, \mathbf Z_{h^i} &= \operatorname{AvgPool}(\mathbf F_{s^i}), \operatorname{AvgPool}(\mathbf F_{h^i}), \\
\mathbf W^C_{s^i},\mathbf W^C_{h^i} &= \operatorname{ChannelAttn}(\mathbf F^i_{sem}, 
\mathbf F^i_{\textit{hfq}}) \\
&= \operatorname{Split}\left(\operatorname{MLP}\left(\left[{\overline{\mathbf F^i_{sem}}}, {\overline{\mathbf F^i_{\textit{hfq}}}}\right]\right)\right),
\end{split}
\end{equation}
where $\operatorname{Split}$ is the reverse operation of $\operatorname{Concat}$. To obtain the spatial attention vector, we use two 1×1 convolutions with an intermediate ReLU layer, denoted by $g(\cdot)$, to aggregate spatial information, spatial-attention vectors $\mathbf W^S_{s^i},\mathbf W^S_{h^i}\in \mathbb{R}^{H_i\times W_i}$ can be obtained:
\begin{equation}
\label{spatial_att}
\begin{split}
\mathbf W^S_{s^i}, \mathbf W^S_{h^i} &= \operatorname{SpatialAttn}\left(\mathbf F^i_{sem}, 
\mathbf F^i_{\textit{hfq}}\right) \\
&= \operatorname{Split}\left(\operatorname{Conv}\left({g}\left(\operatorname{Conv}\left(\left[\mathbf F^i_{sem}, 
\mathbf F^i_{\textit{hfq}}\right]\right)\right)\right)\right).
\end{split}   
\end{equation}

% Finally, we apply the obtained attention vectors to the opposite branch's features and then add them to the current branch's features to obtain the final aligned features, which serve as the input for the next layer, for $i=1,2,3,4$:
Finally, we align branch features by applying attention vectors crosswise, which gives out the input of the next layer by residual adding, for $i=1,2,3,4$:
\begin{equation}
\begin{split}
& \mathbf F^C_{s^i} = \mathbf W^C_{s^i} \odot \mathbf F^{i}_{sem},\quad \mathbf F^S_{s^i} = \mathbf W^S_{s^i} \odot \mathbf F^{i}_{sem}, \\
& \mathbf F^C_{h^i} = \mathbf W^C_{h^i} \odot \mathbf F^{i}_{\textit{hfq}},\quad \mathbf F^S_{h^i} = \mathbf W^S_{h^i} \odot \mathbf F^{i}_{\textit{hfq}}, \\ 
& \mathbf F^{i}_{sem} := \mathbf F^{i}_{sem} + \mathbf F^C_{h^i} + \mathbf F^S_{h^i}, \\
& \mathbf F^{i}_{\textit{hfq}} := \mathbf F^{i}_{\textit{hfq}} + \mathbf F^C_{s^i} + \mathbf F^S_{s^i}.
\end{split}
\end{equation}

% \end{eqnarray}
Then, we fuse semantic feature $\mathbf F^i_{sem}$ and high-frequency feature $\mathbf F^i_{\textit{hfq}}$ to get the input $\mathbf F_d^i$ of the MLN. 
Tampering operations affect a certain number of pixels rather than a single pixel, restricting the attention range is more advantageous in suppressing sporadic positive responses to the features. Therefore, we utilize deformable attention\cite{deformable} for enhancement. 
The fusion process can be described by the following equations:
% The deformable attention mechanism is more efficient in computation. 

\begin{equation}
\label{deform_att}
\begin{split}
&\mathbf {attn}_{s} = \operatorname{DfA}(\operatorname{query}=\mathbf F^{i}_{{sem}}, \operatorname{value}=\mathbf 
 F^{i}_{\textit{hfq}}), \\
&\mathbf {attn}_{h} = \operatorname{DFA}(\operatorname{query}=\mathbf F^{i}_{\textit{hfq}}, \operatorname{value}=\mathbf F^{i+1}_{{sem}}), \\
&\mathbf F_d^i = \gamma_1 * (\mathbf F^{i}_{{sem}} + \mathbf {attn}_{s}) + \gamma_2 * (\mathbf F^{i}_{\textit{hfq}} + \mathbf {attn}_{h}), 
\end{split}
\end{equation}
where $\gamma_1$, $\gamma_2$ are learnable parameters, and DFA means Deformable Attention.

\subsection{Manipulation Localization Network}
MLN adopt the architecture of Mask2Former\cite{Mask2Former}, which involves two parts: the Pixel Decoder and the Transformer Decoder. The Pixel Decoder is primarily responsible for progressively upsampling features from low resolution to high resolution. The Transformer Decoder utilizes a single query embedding and multi-scale features as inputs. The use of multi-scale features is advantageous for locating small tampered regions, while query embeddings, combined with Masked-Attention, help restrict Cross-Attention to the tampered regions for extracting tampering-related features.

\subsection{Prompt Tuning Method} 
We leverage rich semantic features from pre-trained visual foundation models through prompt tuning. For each basic block, we concatenate unique prompt embeddings and image tokens as input:
\begin{eqnarray}
\mathbf F^{i,j} = \operatorname{{Block}^{i,j}_{sem}}\left(\left[\mathbf P^{i,j}, \mathbf F^{i,j-1}\right]\right), i=1,2,...,N_i,
\end{eqnarray}
where $N_i$ symbolizes the total number of blocks in $i$-th layer. 
% We use ``$\underline{\hbox to 4mm{}}$'' as placeholders to indicate that each block will use an independent set of prompt embeddings.
% However, Swin-Transformer's unique window attention mechanism further divides image blocks into $n_w$ windows of size $ws*ws$, calculating self-attention within the windows to reduce time complexity. 
Assume input with batch size of $B$, where $\mathbf P^{i,j} \in \mathbb{R}^{B\times n_p\times C_i}$. We expand $\mathbf P^{i,j}$ after partitioning to alter dimensions to $\mathbb{R}^{(B * n_w)\times n_p\times C}$, ensuring that each window contains exactly $n_p$ prompt embeddings for self-attention computation. After merging windows, we average on $n_w$ groups of prompt tokens to reshape back as $\mathbf P^{i,j} \in \mathbb{R}^{B\times n_p\times C}$.

% \subsection{Objective Loss Function}
% We consider the IML task as a pixel-level binary classification problem, using 1 to represent manipulated pixels and 0 for real pixels. Therefore, we use binary cross-entropy loss as the final loss function. Considering the imbalance between the number of fake pixels and real pixels, we adopt the strategy proposed by [], applying weights to each pixel's loss:
% \begin{gather}
% \mathcal{L} = \frac{1}{N} \sum_{i} - [ w_1 \cdot y_i\cdot \operatorname{log}(\hat{y_i}) + w_0 \cdot (1-y_i)\cdot \operatorname{log}(1 - \hat{y_i})], 
% \end{gather}
% where $N$ is the number of images in a batch, $y_i$, $\hat{y_i}$ represent for groundtruth and prediction respectively, $w_1$, $w_0$ symbolize balancing weights.

\begin{figure}[!t]
\centering
\includegraphics[width=0.49\textwidth]{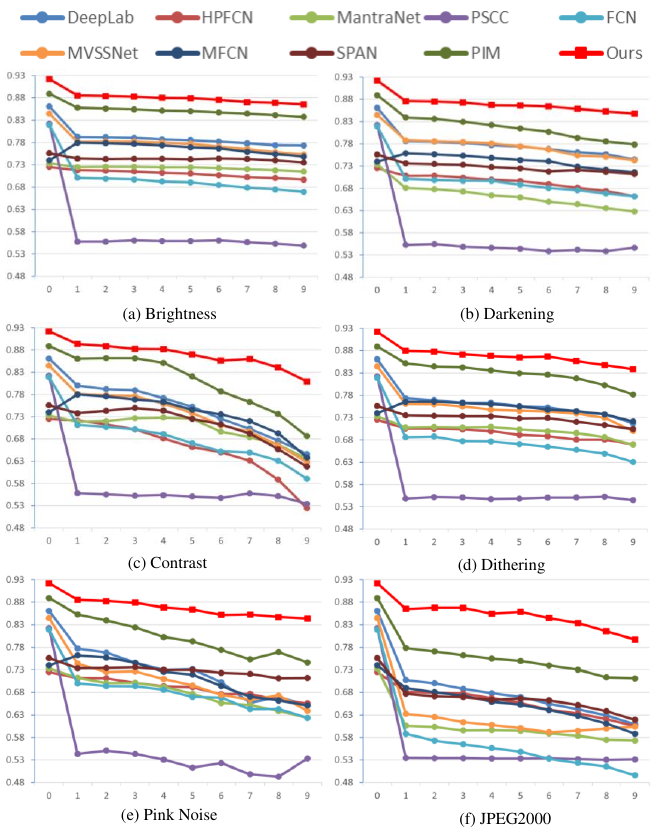} 
\caption{Robustness evaluation against 6 different perturbations. Test dataset is CASIA1, and AUC is the evaluation metric. The x-axis symbolizes the perturbation severity level from 0 to 9 with 0 being no perturbation.} 
\label{rr}
\end{figure}

\begin{table*}[!t]
\small
\begin{center}
\caption{The comparison of image manipulation localization performance (F1 score with fixed threshold: 0.5). The best performance in each column are bolded and the second best underlined.} \label{tab1}
\setlength{\tabcolsep}{2mm}{
\begin{tabular}{cccccccccc}
  \hline
  Method & CASIA1 & NIST16 & COVER & IMD20 & Columbia & DEF-12K & In-the-Wild & Korus & Average
  \\
  \hline
  FCN~\cite{FCN} & 0.441 & 0.167 & 0.199 & 0.210 & 0.223 & 0.130 & 0.192 & 0.122 & 0.211\\
  DeepLabv3~\cite{DeepLabv3} & 0.429 & 0.237 & 0.151 & 0.216 & 0.442 & 0.068 & 0.220 & 0.120 & 0.235\\
  MFCN~\cite{MFCN} & 0.346 & 0.243 & 0.148 & 0.170 & 0.184 & 0.067 & 0.161 & 0.118 & 0.180\\
  RRU-Net~\cite{RRU-Net} & 0.291 & 0.200 & 0.078 & 0.159 & 0.264 & 0.033 & 0.178 & 0.097 & 0.163\\
  HPFCN~\cite{HPFCN} & 0.173 & 0.172 & 0.104 & 0.111 & 0.115 & 0.038 & 0.125 & 0.097 & 0.117\\
  MantraNet~\cite{wu2019mantra} & 0.187 & 0.158 & 0.236 & 0.164 & 0.452 & 0.067 & 0.314 & 0.110 & 0.211\\
  H-LSTM~\cite{H-LSTM} & 0.156 & \textbf{0.357} & 0.163 & 0.202 & 0.149 & 0.059 & 0.173 & 0.143 & 0.175\\
  SPAN~\cite{SPAN} & 0.143 & 0.211 & 0.144 & 0.145 & 0.503 & 0.036 & 0.196 & 0.086 & 0.183\\
  % ViT-B & 0.282 & 0.254 & 0.142 & 0.154 & 0.217 & 0.062 & 0.208 & 0.176 & 0.165\\
  % Swin-ViT & 0.390 & 0.220 & 0.168 & 0.300 & 0.365 & 0.157 & 0.265 & 0.134 & 0.250\\
  PSCC~\cite{PSCC} & 0.335 & 0.173 & 0.220 & 0.197 & 0.503 & 0.072 & 0.303 & 0.114 & 0.240\\
  EVP~\cite{EVP} & 0.483 & 0.210 & 0.114 & 0.233 & 0.277 & 0.090 & 0.231 & 0.113 & 0.219\\
  CAT-Net~\cite{CAT-Net} & 0.237 & 0.102 & 0.210 & 0.257 & 0.206 & \textbf{0.206} & 0.217 & 0.085 & 0.190\\
  MVSS-Net++~\cite{MVSS-Net++} & 0.513 & 0.304 & \textbf{0.482} & 0.270 & 0.660 & 0.095 & 0.295 & 0.102 & 0.340\\
  PIM~\cite{PIM} & \underline{0.566} & 0.280 & 0.251 & \underline{0.419} & \underline{0.680} & 0.167 & \textbf{0.418} & \underline{0.234} & \underline{0.377}\\
  \hline
  Ours & \textbf{0.581} & \underline{0.343} & \underline{0.414} & \textbf{0.423} & \textbf{0.801} & \underline{0.194} & \underline{0.414} & \textbf{0.266} & \textbf{0.430} \\
  \hline
\end{tabular}}
\end{center}
\end{table*}

\section{Experiment}

\subsection{Experimental Setup}
\textbf{Datasets}. During the training phase, we utilize the CASIA2\cite{CASIA2} and synthetic datasets used by PSCC-Net\cite{PSCC} as the training set. To achieve a balance between training effectiveness and efficiency, we randomly sample 30,000 images from the synthetic dataset for training. During the testing phase, we employ eight common datasets to assess our model's performance: CASIA1\cite{CASIA1}, COVER\cite{COVER}, IMD20\cite{IMD20}, NIST16\cite{NIST16}, Columbia\cite{Columbia}, DEFACTO-12K\cite{DEFACTO-12K}, In-the-Wild\cite{In-the-Wild}, and Korus\cite{Korus}. 

% \noindent\textbf{Evaluation Metrics}. The performance of each method will be evaluated by F1 score. We set the fixed threshold of F1 score as 0.5 because the optimal threshold is determined exclusively for cases where ground truth information is available.

\noindent\textbf{Implementation Details}. We train our model on two RTX 3060 GPUs with a batch size of 14. We employ weighted cross-entropy loss as the objective function. Input images are resized to $512\times 512$. We utilize the AdamW optimizer with $\beta_1 = 0.9$, $\beta_2 = 0.999$, weight decay 0.05, and cosine annealing warm restarts strategy. The maximum learning rate is set to 1e-4, and the minimum learning rate is 1e-6. The model is trained for 80 epochs, including 5 warm-up epochs. 

\subsection{Comparisons}
We compare our method with 13 state-of-the-art models to comprehensively assess the model's performance. We follow the metrics of previous work~\cite{PIM} for evaluation, the experimental results are shown in Table~\ref{tab1}, in which the best results are bolded and the second-best results are underlined. 
The proposed model places either 1st or 2nd on all datasets, demonstrating its effectiveness and generalization ability.
% The proposed model achieve the best performance on four datasets and second-best performance on the remaining four datasets, demonstrating good generalization.
It is worth noting that H-LSTM achieves favorable performance on the NIST16 dataset, primarily owing to the specially fine-tuning. 
MVSSNet++ is a meticulously designed network that fully leverages boundary artifacts and high-frequency information from forged images. However, by integrating the semantic and high-frequency information, we achieve a 9\% F1-score improvement in average. 
EVP fails to achieve satisfactory generalization performance, possibly due to its less effective fusion strategy. 
The manipulation localization results of various methods are illustrated in Fig.~\ref {lr}, in which our approach transcends limitations associated with specific types of datasets, demonstrating its efficacy in effectively addressing a wide array of tampering methods.

\begin{table}[!t]
\small
\begin{center}
\caption{Image Manipulation Localization Performance(F1 score with fixed threshold: 0.5)} \label{tab2}
\setlength{\tabcolsep}{2mm}{
\begin{tabular}{cccccc}
  \hline
  Setting & Sem & HP & F.Align & F.Fuse & F1-score \\
  \hline
   1 &\checkmark &- &- &- & 0.481 \\
   2 &- &\checkmark &- &- & 0.392 \\
   3 &\checkmark &\checkmark &- &- &0.505 \\
   4 &\checkmark &\checkmark &\checkmark &- & 0.555\\
   5 &\checkmark &\checkmark &- &\checkmark & 0.517\\
   6 &\checkmark &\checkmark &\checkmark &\checkmark & 0.581 \\
  \hline
\end{tabular}}
\end{center}
\end{table}

\subsection{Robustness Test}
In the real-world scenario, manipulated images may suffer from various post-processing techniques, leading to the fading or disappearance of tampering traces, which significantly compromises the model's performance. 
We follow the setup introduced by \cite{PIM}, introducing six common perturbations to mimic post-processing effects of brightness, contrast, darkening, dithering, pink noise and JPEG2000 compression.
% To assess the capability of our method in handling image post-processing operations, we follow the setup introduced by \cite{PIM}, introducing six common perturbations to mimic post-processing effects, brightness, contrast, darkening, dithering, pink noise and JPEG2000 compression. 
We evaluate the robustness of each method on the CASIA1 dataset, with pixel-level localization AUC scores presented in Fig.~\ref{rr}. 
The results demonstrate the necessity of incorporating semantic information, as the aligned semantic features supplement the high-frequency feature well, which contributes in the robustness of the proposed method.

% The proposed method achieves better robustness primarily because semantic information is less affected by image post-processing operations compared to high-frequency information.

\subsection{Ablation Study}
To assess the effectiveness of the modules we design, we conduct comprehensive ablation experiments. Table \ref{tab2} presents the specific experimental settings and corresponding F1 scores testing on the CASIA1 dataset. Experiment 1 utilize only the semantic branch through prompt tuning. Experiment 2, on the other hand, solely employ a high-frequency branch trained from scratch. The results demonstrate that either semantic or high-frequency information is vital in the IML task.
Furthermore, we investigate the effectiveness of the designed alignment and fusion method via experiments 3 to 5.
We ablate the deformable attention in fusion by substituting it with element-wise addition.
The results exhibit the effectiveness of the designed multiple attention mechanisms.

\section{Conclusions}
We present Prompt-IML, which introduces semantic information of pre-trained visual foundation models into IML tasks. The semantic information is leveraged through prompt tuning and fused with high-frequency information of images. Experimental results on typical IML datasets demonstrate the effectiveness of the proposed method.

\bibliographystyle{IEEEtran}
\small
\bibliography{icme2023template}

\end{document}